\def\year{2022}\relax
\documentclass[letterpaper]{article} 
\usepackage{aaai22}  
\usepackage{times}  
\usepackage{helvet}  
\usepackage{courier}  
\usepackage[hyphens]{url}  
\usepackage{graphicx} 
\usepackage{xcolor}
\usepackage{natbib}  
\usepackage{caption} 
\DeclareCaptionStyle{ruled}{labelfont=normalfont,labelsep=colon,strut=off} 
\usepackage{algorithm}
\usepackage{algorithmic}
\usepackage{amsmath,amssymb}
\usepackage{newfloat}
\usepackage{listings}
\floatstyle{ruled}
\newfloat{listing}{tb}{lst}{}
\floatname{listing}{Listing}

\title{Learning to Minimize Cost-to-Serve for Multi-Node Multi-Product\\Order Fulfilment in Electronic Commerce}
\author{
    Pranavi~Pathakota\equalcontrib\textsuperscript{\rm 1}, Kunwar~Zaid\equalcontrib\textsuperscript{\rm 1}, Anulekha~Dhara\textsuperscript{\rm 1}, Hardik~Meisheri\textsuperscript{\rm 1}, Shaun~D'Souza\textsuperscript{\rm 2}, Dheeraj~Shah\textsuperscript{\rm 2}, Harshad~Khadilkar\textsuperscript{\rm 1}
}
\affiliations{

    \textsuperscript{\rm 1}TCS Research\\
    \textsuperscript{\rm 2}Tata Consultancy Services Limited\\
    India\\
    (p.pranavi,kunwar.zaid)@tcs.com
}

\usepackage{bibentry}

\begin{document}

\maketitle

\begin{abstract}
We describe a novel decision-making problem developed in response to the demands of retail electronic commerce (e-commerce). 
While working with logistics and retail industry business collaborators, we found that the cost of delivery of products from the most opportune node in the supply chain (a quantity called the cost-to-serve or CTS) is a key challenge. The large scale, high stochasticity, and large geographical spread of e-commerce supply chains make this setting ideal for a carefully designed data-driven decision-making algorithm. In this preliminary work, we focus on the specific subproblem of delivering multiple products in arbitrary quantities from any warehouse to multiple customers in each time period. We compare the relative performance and computational efficiency of several baselines, including heuristics and mixed-integer linear programming. We show that a reinforcement learning based algorithm is competitive with these policies, with the potential of efficient scale-up in the real world.
\end{abstract}


\section{Introduction}\label{sec:intro}
Supply chains are complex networks involving the movement of people, activities, information and resources in order to supply a product or service to diverse customers. The management and operation of supply chains has been a problem of interest for decades \cite{holt1955linear,haley1973inventory,lambert2000issues}. The earliest supply chains were focused on movement of material from \textit{resource extraction} to the \textit{manufacturer}. More recent studies considered the `retail' supply chain consisting of material movement from the manufacturer through a \textit{distribution network} (in the form of a hierarchy of warehouses) to retail stores. In the last 20 years, a new paradigm of e-commerce has emerged: that of delivery of goods and products from warehouses (distribution centers) directly to customer doorsteps. This is the version of the supply chain that is the most challenging because of the complex flow of products, and is of interest in the current paper. 

The specific portions of the supply chain corresponding to the e-commerce business involve several sub-problems such as inventory replenishment, job-shop scheduling, bin packing, and vehicle routing.  These individual problems have been of interest to the operations research community for a long time  \cite{ivanov2018survey} and exhibit unique characteristics, posing challenges to smooth and profitable operation \cite{golicic2002impact,lu2015effects}. Approaches for solving these problems include heuristics \cite{cooper64, la20, rabbani21,reph94, reph07, jph06, jph21, vrph13, vrph21} and reinforcement learning \cite{dlla18, ambloc21,reprl20, reprl21, jprl19, jprl20, vrprl18, vrprl21}. 

\textbf{Prior work: }We have been unable to find research studies that formally define and address cost-to-serve (CTS). The problem is typically approached from a process management point of view \cite{rwblog} to recognize that different products and different channels to market having different cost drivers  within the supply chain. The term \textit{cost-to-serve} was used to describe customer-service costs in \cite{cooper97, alan98, robert01}. In a case review of a Swedish manufacturer of heating systems \cite{kaplan89} presented the so called \textit{whale curve}. The analysis of accumulated profitability per customer demonstrated that 20\% of the customers generated 2.25 times the net company profit,  70\%  of  customers  were  on  the  balance  point, and 10\% generated a loss of 1.25 times of net profits. A broader survey of CTS from an operations management perspective is available in \cite{bio08}. Recently, work has been done towards applying CTS in customer selection \cite{li18} and in designing strategies as regards emerging marketing channels \cite{ma15}. CTS has been primarily considered as a management problem and solved using statistics or regression models.

\textbf{Real-world impact: }A significant impact of CTS on the overall supply chain profitability have been studied in \cite{url1, url2}. 
A carefully designed decision-making system should be able to improve CTS and reduce this number, through avenues such as efficient distribution, deferred service, dynamic/personalized pricing, and (in extreme cases) order cancellation. This is a challenging task because of the sequential nature of the problem. Each sourcing, distribution, and pricing decision affects the inventory availability, vehicle capacity, and customer satisfaction, implying that locally greedy policies are unlikely to be optimal. Furthermore, retailers can have different and time-varying objectives based on business goals. All these considerations build a strong case for the use of learning-based techniques in place of the state-of-the-art heuristics used in current real world solutions. 

\begin{figure}[t]
    \centering
    \includegraphics[width=0.47\textwidth]{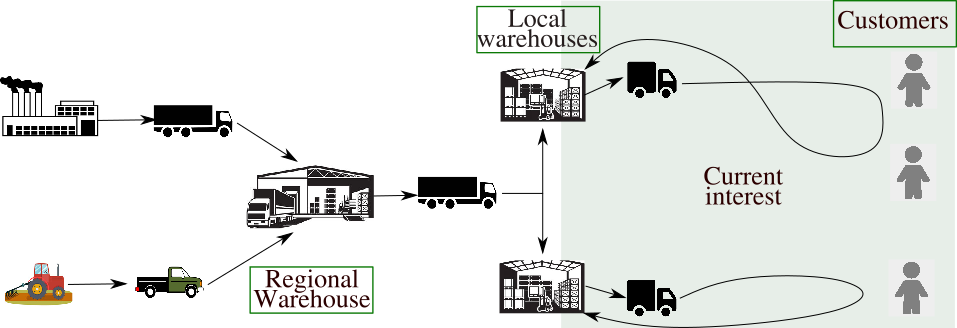}
    \caption{Illustration of e-commerce supply chain. We focus on the portion from local warehouses to the customers.}
    \label{fig:highlevel}
\end{figure}

In this paper, cost-to-serve (CTS) is defined as the cost of delivering multiple products in arbitrary quantities from a warehouse (or multiple warehouses) to customers who have placed orders through the electronic channel (website). We focus specifically on the right half of Figure \ref{fig:highlevel}, leaving out of scope the task of replenishing local warehouses through the upstream supply chain. CTS is used to understand the true cost of distribution, by analysing all activities to reveal the total cost of servicing each individual customer with a specific product. The major contributions of our paper are, (i) a formal definition of cost-to-serve (CTS) as a decision-making problem, (ii) a reinforcement learning (RL) based formulation of a preliminary sub-problem demonstrating the potential for future expansion, (iii) validation and benchmarking of our RL approach against state-of-the-art heuristic policies, and (iv) definition of specific future directions for this work, which will deliver business impact. 

\section{Problem formulation} \label{sec:desc}

We consider only two types of entities in the present work: the distribution centers (warehouses) and customers. At a given time-step, multiple customer orders may be generated for fixed number of products to be delivered from fixed number of warehouses. 
The goal of the algorithm is to fulfil order quantities as much as possible by choosing one warehouse to supply each product in this order. Different products can be supplied from different warehouses if necessary, but the full quantity of a given product must be supplied from a single warehouse. Each warehouse is resupplied (inventory replenished) to its maximum capacity periodically, independently of dynamic demand. Each customer order must be fulfilled in the current time step if sufficient inventory is available (deferred fulfilment is something we are currently working on). Any orders that are not available in the warehouses are considered to have been dropped. 



Assume that the distribution network consists of $N$ warehouses with each warehouse $j=\{1,2,\ldots,N\}$ stocking $i=\{1,2,\ldots,M\}$ product types up to a maximum level $p_{ij}^\mathrm{max}$. Denote the $j^\mathrm{th}$ warehouse location by $w_j = (w_x^j,w_y^j)$. At time step $t$, let $p_{ij}^{(t)}\in\mathbb{Z}^+$ denote the quantity available for the $i^\mathrm{th}$ product at $j^\mathrm{th}$ warehouse, and let $d_{ik}^{(t)}\in\mathbb{Z}^+$ denote the demand for the $i^\mathrm{th}$ product by customer $k$ in that time step (where $\mathbb{Z}^+$ is the set of non-negative integers). Let the location of the customer $k$ at time $t$ be $c_k^{(t)} = (c_{k,x}^{(t)},c_{k,y}^{(t)})$, and the distance from this customer to warehouse $j$ be $d(w_j,c_k^{(t)})$. Further assume that all products in all warehouses are stocked to their maximum levels $p_{ij}^\mathrm{max}$ after every $T$ time steps. 

At time step $t$, a warehouse $j$ is chosen to supply the requested quantity $d_{ik}^{(t)}$ of each product $i$ for customer $k$. The order is then packed in cartons based on a \textit{quantization number} $q_i$ for each product, which is the number of items of product $i$ that can be packed into a single carton. Each product can have a different quantization number. The total number of cartons/boxes for the $i^\mathrm{th}$ product of the $k^\mathrm{th}$ customer to be delivered from the $j^\mathrm{th}$ warehouse is given by,
\begin{equation*}
    b_{ijk}^{(t)} = \begin{cases} \mathcal{d}(d_{ik}^{(t)}/q_i)\mathcal{e} & \text{ if }i\text{ delivered from }j \\ 0 & \text{otherwise}. \end{cases}
\end{equation*}

We assume that different products are not combined into the same cartons. With this background, we can define the cost components of CTS for the present study as follows:

\begin{enumerate}
    \item Transportation cost, proportional to the distance between the various chosen warehouses and the customer location. This component is independent of the number of cartons, and only counts `trips' made by vehicles. 
    
    \item Carton cost, proportional to number of cartons required.
    
    \item Warehouse cost: This is a fixed cost applied for each warehouse that sources at least one product in a given order, and can vary from one warehouse to another.
    
\end{enumerate}


\section{Methodology} \label{sec:experiment}

\subsection{Baselines} \label{subsec:baseline}

We could not find any explicitly defined heuristics for CTS in literature, and we therefore chose to adapt heuristics defined for the location allocation problem (see literature review). We use the following exact optimization and heuristics as baselines: 
\begin{enumerate}
    \item An exact mixed-integer linear program with the objective of minimizing the distance, carton and warehouse costs while fulfilling as many products as possible.
    \item A 1-step greedy policy $\pi_{1}$ 
    is a heuristic that chooses the nearest feasible warehouse to fulfill the orders. 
    \item A more sophisticated 2-step greedy policy $\pi_{2}$. 
    This heuristic attempts to fulfil the entire order from the second-nearest warehouse if the nearest one cannot do so; this reduces the `warehouse cost' term defined earlier. In the present definition of the problem, $\pi_{2}$ is optimal in the single-time-step sense provided the inventory levels are same. It only leaves room for improvement if another policy has the foresight to accept suboptimal decisions for the short term in order to retain better inventory levels for the longer term.
   \item Enhanced versions $\pi_{P1}$ and $\pi_{P2}$ of policies $\pi_1$ and $\pi_2$ are also considered by prioritising the customers arriving in the same time-step based on the smallest distance to the warehouses. In experiments, we observe that these policies always outperform the original policies. Moreover, for single customer case (1-customer arriving at a time-step), $\pi_{P1}$ and $\pi_{P2}$ coincide with $\pi_1$ and $\pi_2$ respectively. 
\end{enumerate}

\subsection{Reinforcement learning solution} \label{subsec:states}

\begin{figure}[t]
\centering
\includegraphics[width=0.8\linewidth]{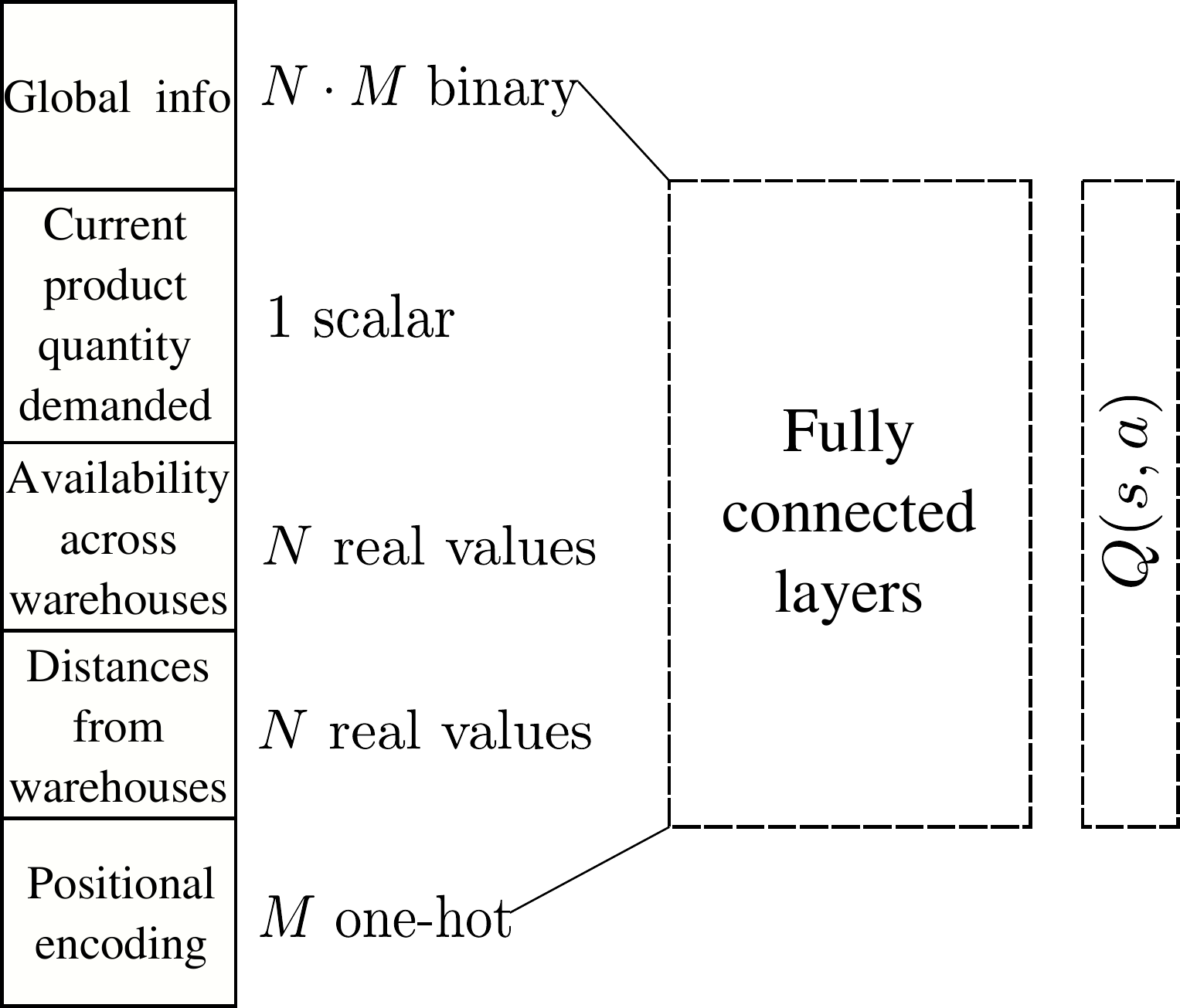}
\caption{RL agent architecture for each product.}
\label{fig:RL_arch}
\end{figure}

Our RL approach is based on the Deep Q network (DQN)~\cite{mnih2015human}. The decision variable (action) in our problem is to select which warehouse each product should be shipped from for each customer. For a single customer at timestep $t$, If we consider this as a concurrent decision in the traditional RL framework, the action space quickly becomes intractable ($N \times M$ sized). Furthermore, such an action space would vary with the number of products, which is undesirable in a business where new products are introduced (and old ones discontinued) very frequently. 

To mitigate this issue, we have broken down this problem into decisions for a single product, which are then concatenated to complete the decision-making step. For each product, we use the network depicted in Figure \ref{fig:RL_arch} to generate decisions. The forward pass through this network is done in parallel (time distributed layers) with weights being shared across the products. During backpropagation, gradients from all these decisions are averaged out to update weights. Note that the weight sharing and parallel execution allows us to seamlessly change the number and variety of products ordered by each customer. However, the combined decisions drive the total cost (based on the number of activated warehouses and other quantities), thus requiring some amount of coordination among the products. This is handled by the state-space representation.

\begin{table}[t]
\centering
\begin{tabular}{|p{2.0cm}|p{3.5cm}|p{1.0cm}|}
\hline
\textbf{Feature}             & \textbf{Description}                                                                                               & \textbf{Size}         \\ \hline
Distance            & Distance $d(w_j,c^{(t)})$ from each warehouse                                          & $N$          \\ \hline
Local Availability  & Quantity $p_{ij}^{(t)}$ of current product available at each warehouse                                                   & $N$          \\ \hline
Demand              & Customer demand $d_i^{(t)}$ for current product                                                              & 1            \\ \hline
Positional Encoding & One-hot encoding of product, with 1 in the $i^\mathrm{th}$ position & $M$          \\ \hline
Global Info & Binary matrix $A$, entries $a_{i,j} = 1$ iff $p_{ij}^{(t)} \geq d_i^{(t)}$ & $N \times M$ \\ \hline
\end{tabular}%
\caption{Features in state representation for each product $i$.}
\label{tab:statespace}
\end{table}

We define a state input that contains both local and global information relative to the current product. Table~\ref{tab:statespace} shows the features used to represent state consisting of both local as well as global information to ensure decision at product level has sufficient context. The global availability matrix provides each product with an indication of which warehouses can supply all the other products, as does the distance vector from all warehouses to the current customer. These inputs help the algorithm reduce the number of activated warehouses and to choose ones that are close to the customer location. The local availability vector helps the algorithm anticipate rewards for future orders of the current product. 

Rewards for the individual products are based directly on the costs previously defined, with the sole difference being that the transportation cost used for training is only from the selected warehouse for product $i$. The remaining terms are given in full to each product (as negative rewards). For $k$ customers, we repeat the above inference process after deducting the demand quantities that we fulfilled from previously served customers. We assume the priority customers to be given, or generated on basis of heuristic.  

During training, we observed that DQN quickly learns to fulfil all the orders, but is unable to learn nuances such as tradeoffs between order splitting and greedy warehouse selection. In order to provide this ability, we borrow an idea from the safe RL domain, where agents are made to respect safety constraints during training as well as deployment. To accelerate learning and avoid catastrophic scenario (not able to fulfil the customer order), we mask the actions which are infeasible, that is having demand which cannot be fulfilled by any warehouse. This has been shown to improve RL performance in safe RL and real-world applications~\cite{garcia2015comprehensive}. We use this mask during the $\epsilon-$greedy training as well, because random exploration does not provide any guarantees of safety~\cite{leike2017ai, ecoffet2019go} and might lead to biased samples in experience replay. 


We have used 4 hidden layers ([40, 20, 20, 20] neurons) neural network for DQN, with $\mathtt{tanh}$ activation in hidden layers and linear activation in the output layer. The training was done using $\mathtt{adam}$ optimizer with a learning rate of $0.001$ and 128 batch size.


\section{Results}

\subsection{Experimentation}


Throughout our experiments, we have considered a 2D grid environment with $N = 4$  warehouses, $M=10$ products at each time step $t$. Warehouse location were fixed at the centre of each quadrant. Customer locations in each time step were chosen with uniform random probability over the $x$ and $y$ axes. The underlying demand distribution for every product was drawn randomly from the uniform, gamma, exponential, normal and poison distributions and then used to generate demand for the entire episode. Replenishment of all products and warehouses happened after $T=50$ time steps, to their maximum capacity values (which are randomly generated for each product). 

We show experiments with up to $k$-customers arriving in a single time step, where $k=1, 4, 10, 40, 80, 100$. The per-unit weightage for each of the cost components is as follows: the distance cost is 100, the carton cost is 5 and the warehouse cost is 40. The main objective of the baselines is to fulfil as many products as possible in the customer order while minimizing the total cost at that time-step. However, an unfulfilment penalty of 100 towards each product which is not fulfilled is  considered in the overall cost while training the RL policy. 
We train the RL agent purely on a $k=1$ or single-customer environment, and test the trained policy on separate data sets created for $k=1, 4, 10, 40, 80, 100$.
\begin{figure}
\centering
\includegraphics[width=0.9\linewidth]{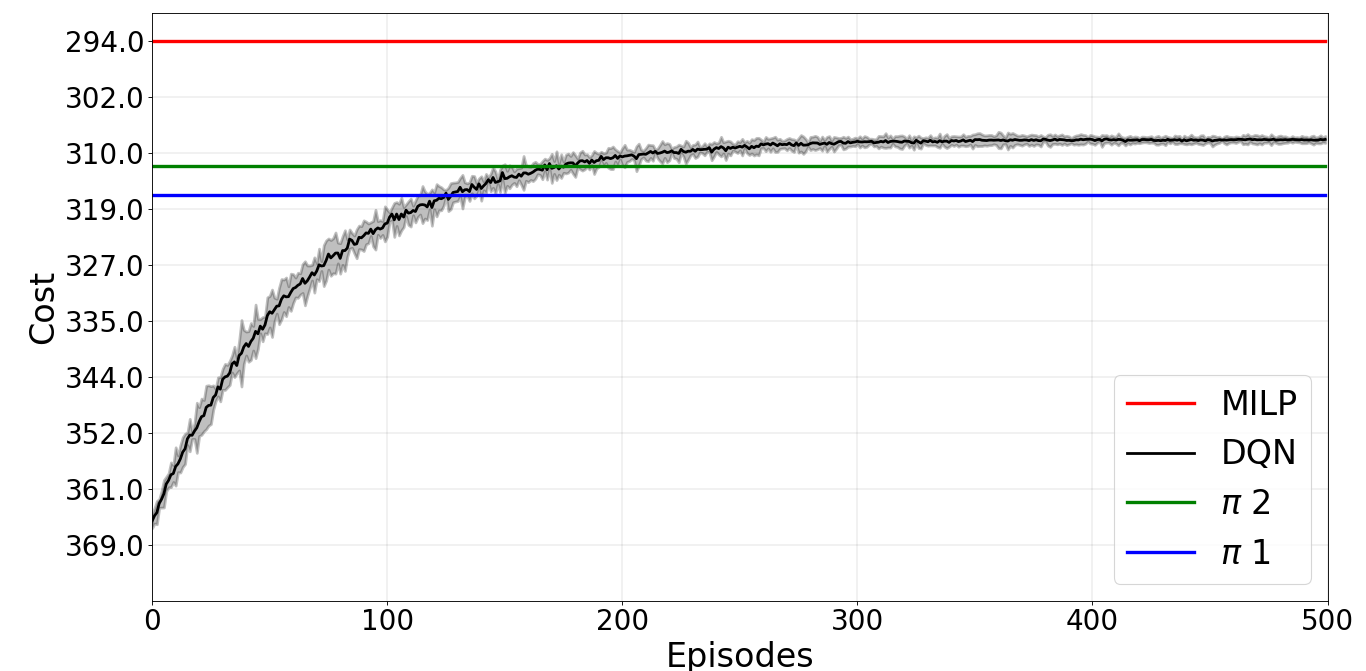}
\caption{Training Plot for DQN.}
\label{fig:dqn_training}
\end{figure}

Figure~\ref{fig:dqn_training} shows the training of the RL in comparison to the performance of the baselines for the 1-customer case. The unfulfilment penalty is added to the total cost of the baselines for fair cost comparison.

\subsection{Comparison with baselines on test data}

\begin{figure}
\centering
\includegraphics[width=0.99\linewidth]{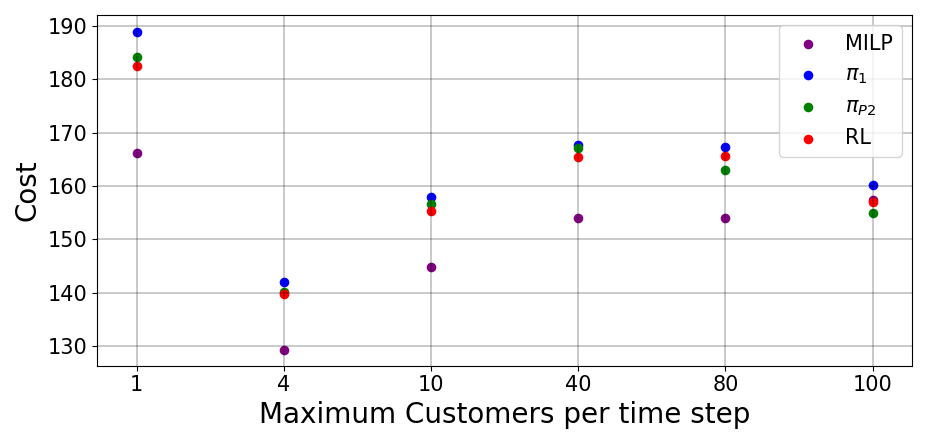}
\caption{Cost comparison between MILP, $\pi_1$, $\pi_{P2}$ and RL.}
\label{fig:results_testing2}
\end{figure}

Figure~\ref{fig:results_testing2} shows total cost comparison (without the unfulfilment penalty) of the RL algorithm with different baselines namely MILP, $\pi_1$ (worst performing policy) and $\pi_{P2}$ (best performing policy) on all 6 test data sets. We observe that RL with DQN is able to outperform $\pi_{P2}$ when maximum number of customers $k$ is less than or equal to 40. RL models are only trained for 1-customer case and trained models are used for inference for $k>1$. RL seems to struggle when $k$ values are increasing as this corresponds to out of distribution states for the model. For 100-customer case, $\pi_{P2}$ and RL perform better than the MILP with respect to the total cost but the number of unfulfilled products for MILP is far less than the other baselines as presented in Table~\ref{tab:statespace}.

\begin{table}[h]
\resizebox{\linewidth}{!}{%
\centering
{\footnotesize{\begin{tabular}{|c||c|c|c|c|c|c|}
\hline 
Average & MILP & $\pi_1$  & $\pi_2$ & $\pi_{P1}$ & $\pi_{P2}$ & RL \\ [0.2ex]\hline\hline
{$\mathbf{1}${-cust}}$_{tt}$ & {0.0985} & {0.0001} & {0.0003} & {0.0002} & {0.0003} & {0.0051} \\ [0.2ex]
{$\mathbf{1}${-cust}$_{up}$}  & 1.273 & 1.277 & 1.282 & 1.277 & 1.282 & 1.2774 \\ [0.2ex]\hline
{$\mathbf{4}${-cust}}$_{tt}$ & 0.1387 &	0.0002 & 0.0004 & 0.0004 &	0.0007 & 0.0120 \\ [0.2ex]
{$\mathbf{4}${-cust}$_{up}$}  & 1.479 & 1.479 & 1.486 & 1.484 & 1.493 & 1.494 \\ [0.2ex] \hline
{$\mathbf{10}${-cust}}$_{tt}$ & 0.3424 & 0.0004 & 0.0010 & 0.0016 &	0.0030 & 0.0264 \\ [0.2ex]
{$\mathbf{10}${-cust}$_{up}$}  & 5.535 & 5.586 & 5.595 & 5.565 & 5.571 & 5.5572 \\ [0.2ex] \hline
{$\mathbf{40}${-cust}}$_{tt}$ & {0.7815} & {0.0011} &{0.0027} & {0.0144} & {0.0325} & {0.0949} \\ [0.2ex]
{$\mathbf{40}${-cust}$_{up}$}  & 25.514 & 25.827 & 25.835 & 25.779 & 25.777 & 25.757 \\ [0.2ex] \hline
{$\mathbf{80}${-cust}}$_{tt}$ & 2.2257 & 0.0025 & 0.0075 & 0.0651 &	0.1460 & 0.2010 \\ [0.2ex]
{$\mathbf{80}${-cust}$_{up}$}  & 55.272 & 63.818 & 63.821 & 63.601 & 63.7 & 63.680 \\ [0.2ex] \hline
{$\mathbf{100}${-cust}}$_{tt}$ & 4.0623 & 0.0033 & 0.0077 &	0.1148 & 0.2325 & 0.2479 \\ [0.2ex]
{$\mathbf{100}${-cust}$_{up}$}  & 72.426 & 104.64 & 104.641 & 104.399 & 104.524 & 104.4674 \\ [0.2ex] 
\hline
\end{tabular}}}%
}
\caption{Average time-taken (tt) in sec and number of unfulfilled products (up) for 6 test data sets with increasing number of customers able to arrive in a single time step.}
\label{tab:statespace}
\end{table}

We note from Table \ref{tab:statespace} that as $k$ increases, RL and $\pi_{P2}$ (the two best-performing policies apart from MILP) have a significantly higher average count of unfulfilled orders. On the other hand, they have a much larger advantage (proportionally speaking) in terms of computation time over MILP. For a system that is designed for real-time decisions with very high volume, this consideration is likely to be the key constraint in the problem. Furthermore, while $\pi_{P2}$ is competitive in this relatively simple environment, the real-world system with much higher complexity will not be trivial to resolve using hand-written rules. Therefore, we believe RL to be the most promising approach for the real-world deployment with all its scale and complexity.




\section{Conclusion}
In this preliminary study, we solve a simplified version of CTS with point decisions for warehouse selection, for multiple customers using RL and several baslines. For now we have focused on transportation and packaging cost and on maximum order fulfilment at every time step.  We are working on incorporating several realistic extensions to the present work. The idea is strongly motivated by recent real-world developments, especially the need for real-time decisions on stocking, order handling and transportation to drive efficiency in a low-margin business. A dynamic decision-making algorithm will not only help in computing the real-time CTS for a customer order, but will also provide a range of alternatives balancing cost and timeliness, with the view of maximising satisfaction and minimising attrition. 

\bibliography{refs}

\end{document}